\documentclass[10pt,twocolumn,letterpaper]{article}

\usepackage{wacv}              

\definecolor{wacvblue}{rgb}{0.21,0.49,0.74}
\usepackage[pagebackref,breaklinks,colorlinks,allcolors=wacvblue]{hyperref}
\usepackage{multirow}
\usepackage[table]{xcolor}

\def\wacvPaperID{254} 
\def\confName{WACV}
\def\confYear{2027}

\usepackage{xcolor}

\definecolor{CrossA}{HTML}{5B2C83}
\definecolor{CrossB}{HTML}{6F3FA3}
\definecolor{CrossC}{HTML}{8453BA}
\definecolor{CrossD}{HTML}{9A68CF}
\definecolor{CrossE}{HTML}{AE7FE0}

\definecolor{WeaverA}{HTML}{C55A11}
\definecolor{WeaverB}{HTML}{D66A1F}
\definecolor{WeaverC}{HTML}{E57A2C}
\definecolor{WeaverD}{HTML}{F08A3A}
\definecolor{WeaverE}{HTML}{F5A15A}
\definecolor{WeaverF}{HTML}{F7B06D}

\title{
{\color{CrossA}C}{\color{CrossB}r}{\color{CrossC}o}{\color{CrossD}s}{\color{CrossE}s}%
{\color{WeaverA}W}{\color{WeaverB}e}{\color{WeaverC}a}{\color{WeaverD}v}{\color{WeaverE}e}{\color{WeaverF}r}:
Cross-modal Weaving for Arbitrary-Modality Semantic Segmentation
}

\author{
Zelin Zhang\\
The University of Sydney\\
{\tt\small zzha0083@uni.sydney.edu.au}
\and
Kedi Li\\
The University of Sydney\\
{\tt\small keli0203@uni.sydney.edu.au}
\and
Huiqi Liang\\
The University of Sydney\\
{\tt\small hlia0544@uni.sydney.edu.au}
\and
Chuanzhi Xu\\
The University of Sydney\\
{\tt\small chuanzhi.xu@sydney.edu.au}
\and
Tao Zhang\\
University of Technology Sydney\\
{\tt\small tao.zhang-8@student.uts.edu.au}
\and
Weidong Cai\\
The University of Sydney\\
{\tt\small tom.cai@sydney.edu.au}
}

\newcommand{\maketeaserfigure}{%
 \vspace{-6pt}
 \noindent\begin{minipage}{\textwidth}
   \centering
   \includegraphics[width=\linewidth]{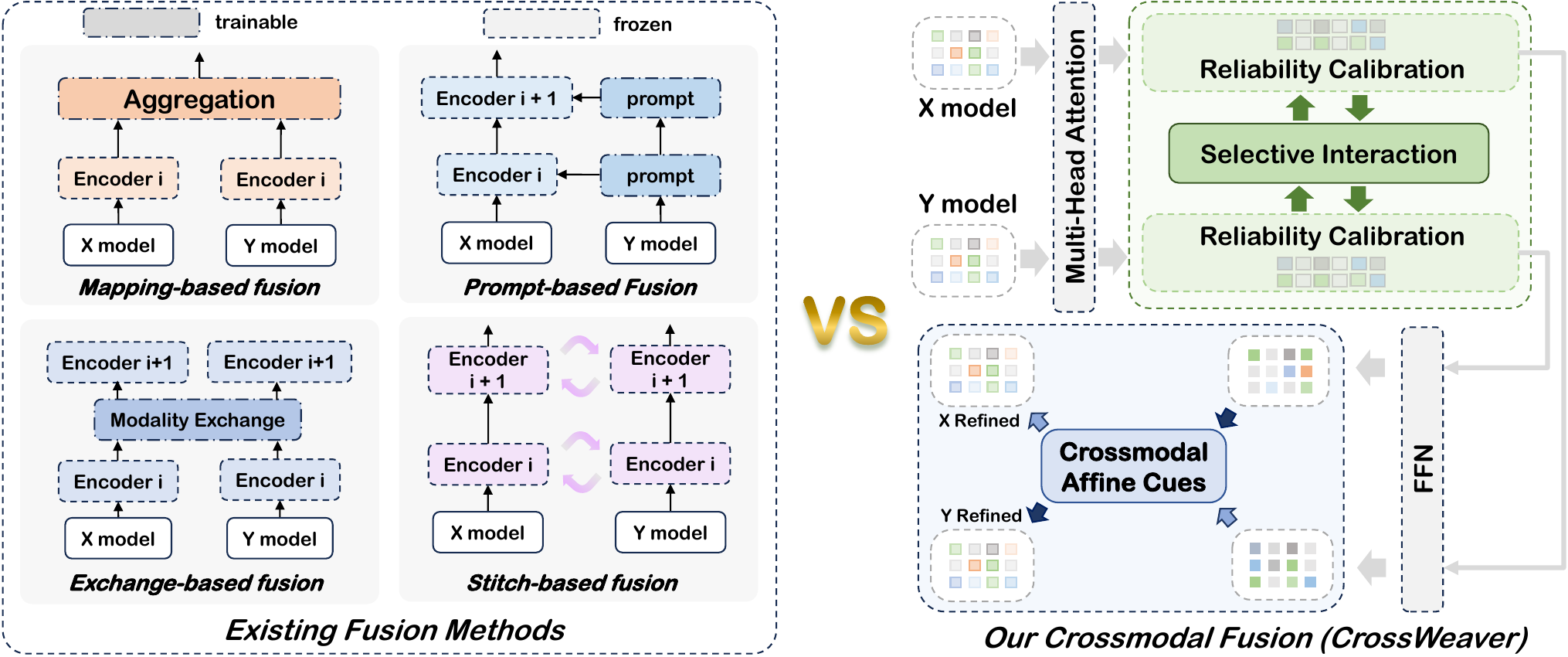}
   \captionsetup{hypcap=false}
   \captionof{figure}{\textbf{Existing multimodal fusion paradigms versus the proposed CrossWeaver.} CrossWeaver enables selective, reliability-aware cross-modal interaction instead of fixed or uniform fusion.}
   \label{fig:overview}
 \end{minipage}
 \vspace{8pt}
}

\makeatletter
\g@addto@macro\@maketitle{\maketeaserfigure}
\makeatother
\begin{document}
\maketitle

\begin{abstract}
Multimodal semantic segmentation has shown great potential in leveraging complementary information across diverse sensing modalities. However, existing approaches often lean upon carefully designed fusion strategies that either use modality-specific adaptations or lean upon loosely coupled interactions, thereby limiting flexibility and resulting in less effective cross-modal coordination. Moreover, these methods often struggle to distinguish reliable complementary cues from noisy or redundant information across different modality combinations. To address these challenges, we propose CrossWeaver, a reliability-aware multimodal interaction framework for arbitrary-modality semantic segmentation. Its core is a Modality Interaction Block (MIB), which enables selective and token-adaptive cross-modal interaction within the encoder, while a Seam-Aligned Fusion (SAF) module further aggregates the enhanced features with improved spatial coherence. Extensive experiments on MCubeS, DeLiVER, and MUSES demonstrate that CrossWeaver achieves state-of-the-art performance and strong generalization to unseen modality combinations.
\end{abstract}

\section{Introduction}
\label{sec:introduction}

Semantic segmentation plays a fundamental role in scene understanding \cite{Zhao2017PSPNetCVPR,zheng2024centering}, autonomous driving \cite{cordts2016cityscapes,Chen2018DeepLabTPAMI}, and robotic perception \cite{FCN,everingham2015pascal} by predicting per-pixel semantic labels of visual inputs. It serves as a cornerstone task for high-level vision applications, including obstacle detection, environment mapping, and urban scene parsing \cite{vobecky2025unsupervised,arulananth2024semantic}. With the rise of deep neural networks, single modality segmentation models such as FCN \cite{FCN} and DeepLab \cite{Chen2018DeepLabTPAMI} have achieved impressive performance on large-scale RGB benchmarks. However, these methods are inherently limited by the information content of a single modality. Relying solely on visual appearance cues is often unreliable under adverse conditions, such as poor illumination, fog, motion blur, or occlusion \cite{zhang2023cmnext,zhang2023cmx}. To overcome these limitations, recent research has increasingly focused on multimodal semantic segmentation \cite{hazirbas2016fusenet,park2017rdfnet,li2025stitchfusion,zheng2024centering}, which leverages complementary sensing modalities such as depth, LiDAR, event, and infrared (IR) to enrich visual understanding.

\begin{figure}[t]
  \includegraphics[width=\linewidth]{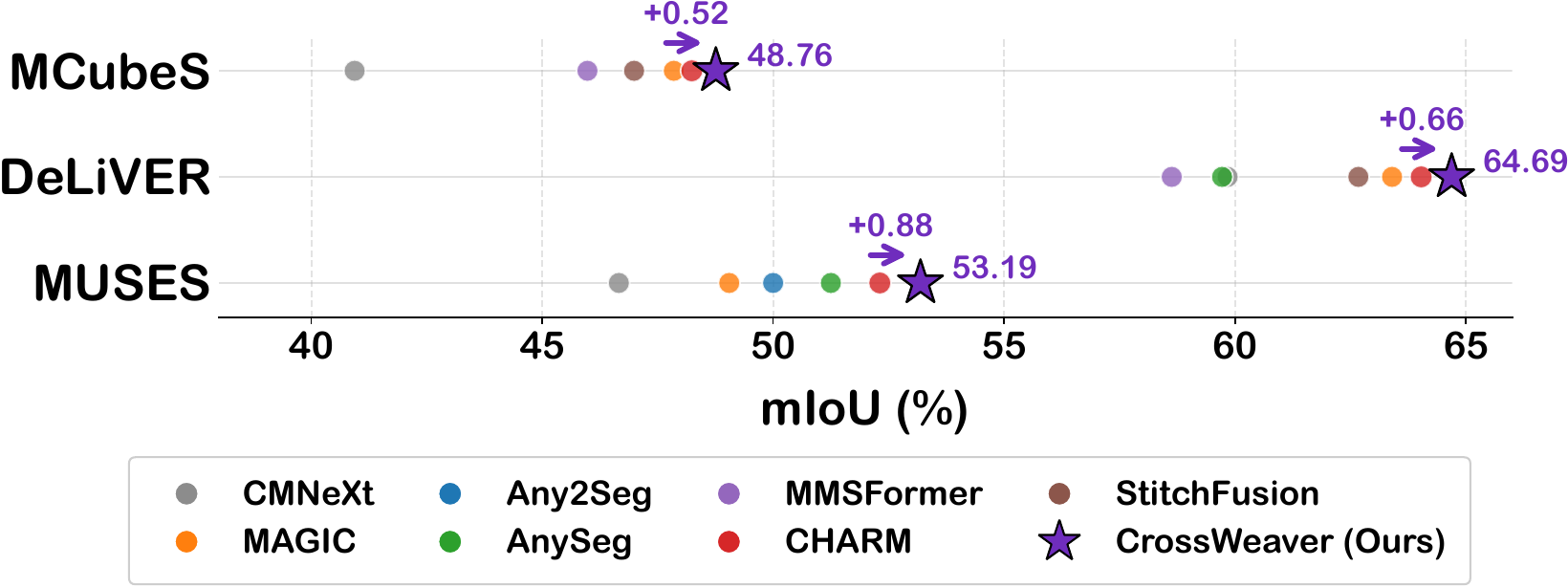}
  \caption{\textbf{Comparison of representative state-of-the-art methods} on MCubeS \cite{liang2022mcubes}, DeLiVER \cite{zhang2023cmnext}, and MUSES \cite{muses}. CrossWeaver is highlighted with purple stars, and purple arrows indicate the improvement over the strongest available baseline on each dataset.
}
  \vspace{-10pt}
  \label{fig:comparison}
\end{figure}

Different modalities provide complementary cues for scene understanding. For example, RGB captures appearance and texture, depth and LiDAR provide geometric structure, while event or infrared signals can offer additional information under challenging lighting or dynamic conditions \cite{hazirbas2016fusenet,park2017rdfnet}. By integrating these heterogeneous signals, multimodal systems can achieve stronger generalization and robustness under complex scenes and diverse environmental conditions, particularly in safety-critical applications such as autonomous driving, where sensor degradation and environmental variation are inevitable.

However, despite these advantages, existing multimodal segmentation frameworks still face several challenges under arbitrary-modality settings. As illustrated in Fig.~\ref{fig:overview}, prior methods have explored different fusion strategies, including mapping-based aggregation, prompt-based adaptation, token exchange, and stitch-based feature fusion. These approaches provide valuable foundations for multimodal representation learning, but they often rely on relatively fixed interaction patterns or uniform information exchange across modalities. As a result, they may be less effective when modality reliability varies across scenes or when only a subset of sensors is available at inference time.

This motivates us to focus on a more systematic reliability-aware interaction design for arbitrary-modality segmentation. In real multimodal perception, different modalities are not equally informative for every scene, region, or semantic class. Some modalities may provide useful complementary cues, while others may be noisy, misaligned, or unavailable. Therefore, effective multimodal interaction should not simply inject or aggregate all modality features uniformly. Instead, it should selectively determine which cross-modal cues are reliable, where they should be introduced, and how they should be integrated into each target representation.

Many existing frameworks also remain RGB-centric, where the fusion process is dominated by the RGB stream while other modalities mainly serve as auxiliary cues \cite{zhang2023cmx,zhang2023cmnext,reza2024mmsformer}. Such an asymmetric design limits flexibility when adapting to diverse or missing modality combinations. Moreover, inherent inconsistencies among modalities, including spatial misalignment, asynchronous sensing, and heterogeneous feature distributions, may further degrade segmentation accuracy near object boundaries \cite{li2025stitchfusion,hazirbas2016fusenet}. These issues motivate CrossWeaver to model cross-modal interaction as a selective and reliability-aware process, rather than as uniform feature injection.

To tackle these challenges, we propose \textbf{\underline{Cross}-modal \underline{Weaver} (\underline{CrossWeaver})}, a reliability-aware multimodal interaction framework that selectively exploits complementary information from different modalities for arbitrary-modality semantic segmentation, as illustrated in Figure~\ref{fig:overview}. 
The core of CrossWeaver is the \textbf{\underline{M}odality \underline{I}nteraction \underline{B}lock (\underline{MIB})}, which enables reliability-aware and token-adaptive cross-modal interaction within a shared hierarchical encoder. Rather than uniformly absorbing information from all modalities, MIB identifies reliable and informative cues from available modalities, allowing each target token to benefit from useful cross-modal context while reducing the influence of noisy, redundant, or less relevant signals. 
Built upon this interaction mechanism, we further employ a \textbf{\underline{S}eam-\underline{A}ligned \underline{F}usion (\underline{SAF})} module to aggregate interaction-enhanced features at each stage, improving spatial coherence while preserving fine-grained modality-specific details. 
This design allows CrossWeaver to better exploit modality complementarity under different sensor combinations, leading to robust and consistent segmentation performance even when some modalities are degraded or missing.

We extensively evaluate CrossWeaver on three multimodal semantic segmentation benchmarks, including MCubeS, DeLiVER, and MUSES. Under the main comparison setting, CrossWeaver achieves 48.76\%, 64.69\%, and 53.19\% mIoU on the three datasets, respectively. As shown in Fig.~\ref{fig:comparison}, CrossWeaver consistently performs favorably against recent state-of-the-art methods and improves over the strongest available baseline on each benchmark. Moreover, when trained with all available modalities and evaluated on arbitrary modality subsets, CrossWeaver attains 32.68 mean mIoU on MCubeS and 40.52 mean mIoU on DeLiVER, highlighting its robustness to missing modalities.

Our main contributions are summarized as follows:
\begin{itemize}
  \item We propose \textbf{CrossWeaver}, a reliability-aware multimodal interaction framework designed for arbitrary modality settings.
  \item We introduce the \textbf{Modality Interaction Block (MIB)}, the core component of CrossWeaver, which enables selective cross-modal interaction by jointly modeling modality reliability, token importance, and interaction consistency within the encoder.
  \item We further develop a lightweight \textbf{Seam-Aligned Fusion (SAF)} module to aggregate interaction-enhanced features with improved spatial coherence and boundary consistency.
  \item Extensive experiments on MCubeS, DeLiVER and MUSES demonstrate that CrossWeaver achieves \textbf{new state-of-the-art mIoU of 48.76\%, 64.69\% and 53.19\%}, respectively, while showing strong robustness under arbitrary modality combinations.
\end{itemize}


\section{Related Work}
\label{sec:related_work}

\noindent\textbf{Semantic Segmentation.}
Semantic segmentation aims to assign a semantic label to each pixel and has been widely studied in scene understanding, autonomous driving, and robotic perception. Since Fully Convolutional Networks (FCNs) \cite{FCN}, representative methods such as DeepLab \cite{Chen2018DeepLabTPAMI,chen2018encoder} and PSPNet \cite{zhao2017pyramid} have improved dense prediction through atrous convolution, encoder-decoder refinement, and pyramid context aggregation. Subsequent works further enhance segmentation with multi-scale feature learning \cite{chen2017rethinking,Hou2020StripPoolingCVPR}, attention mechanisms \cite{fu2019dual,huang2019ccnet}, edge-aware refinement \cite{yu2018bisenet}, transformer architectures \cite{zheng2021rethinking,xie2021segformer,zhang2025efficient}, and alternative token-mixing designs \cite{yu2022metaformer,yuan2018ocnet,liu2022convnet,dong2023egfnet}. Despite these advances, RGB-based segmentation remains vulnerable under challenging conditions such as poor illumination, motion blur, fog, and texture ambiguity \cite{li2025stitchfusion,dong2024understanding}, motivating the use of complementary sensing modalities.

\noindent\textbf{Multimodal Fusion.}
Multimodal fusion improves visual perception by integrating heterogeneous signals such as depth, thermal, polarization, event streams, LiDAR, and optical flow \cite{hazirbas2016fusenet,wang2018depth,dong2024efficient,zhou2022edge,zhang2023cmx,liu2025sharecmp,zhang2023cmnext,chaoyievent,zhu2021cylindrical,yin2023dformer,liu2020afnet,sun2019rtfnet}. Beyond semantic segmentation, it has also been explored in object detection \cite{liu2022target,cao2024bi}, medical imaging \cite{ronneberger2015u,zhou2024multimodal}, and optical flow estimation \cite{zhao2023flowtext,li2024u3m}. Existing strategies range from early concatenation and feature summation to attention-based interaction and dynamic selection \cite{wang2022multimodal,he2024prompting}. While effective, many designs still rely on fixed rules, modality-specific branches, or uniform information exchange, which may introduce redundancy and limit adaptability when sensors are missing, degraded, or unreliable \cite{jia2024geminifusion,dong2024efficient}.

\noindent\textbf{Multimodal Semantic Segmentation.}
For semantic segmentation, early RGB-D methods such as FuseNet \cite{hazirbas2016fusenet} and RDFNet \cite{park2017rdfnet} demonstrated the benefit of combining geometric and appearance cues. Recent methods, including CMX \cite{zhang2023cmx} and CMNeXt \cite{dong2024efficient}, extend multimodal segmentation to more flexible modality settings through cross-modal interaction and token exchange, while other works investigate stronger feature aggregation, modality-specific adaptation, and robust fusion under incomplete inputs \cite{reza2024mmsformer,zhang2024electrical,yang2024polymax,zheng2024centering,li2025stitchfusion,egformer}. However, it remains challenging to exploit complementary information across scales \cite{zhang2025multi,zhang2021abmdrnet}, maintain cross-modal consistency \cite{seichter2022efficient,lv2024context}, and balance accuracy with efficiency \cite{li2024u3m,zhou2022edge}, especially when modalities are noisy, misaligned, or partially unavailable \cite{jia2024geminifusion,dong2024efficient}. CrossWeaver addresses these issues through reliability-aware encoder interaction and seam-aligned fusion.

\section{Methodology}
\label{sec:method}

\begin{figure*}[t]
  \centering
  \includegraphics[width=\linewidth]{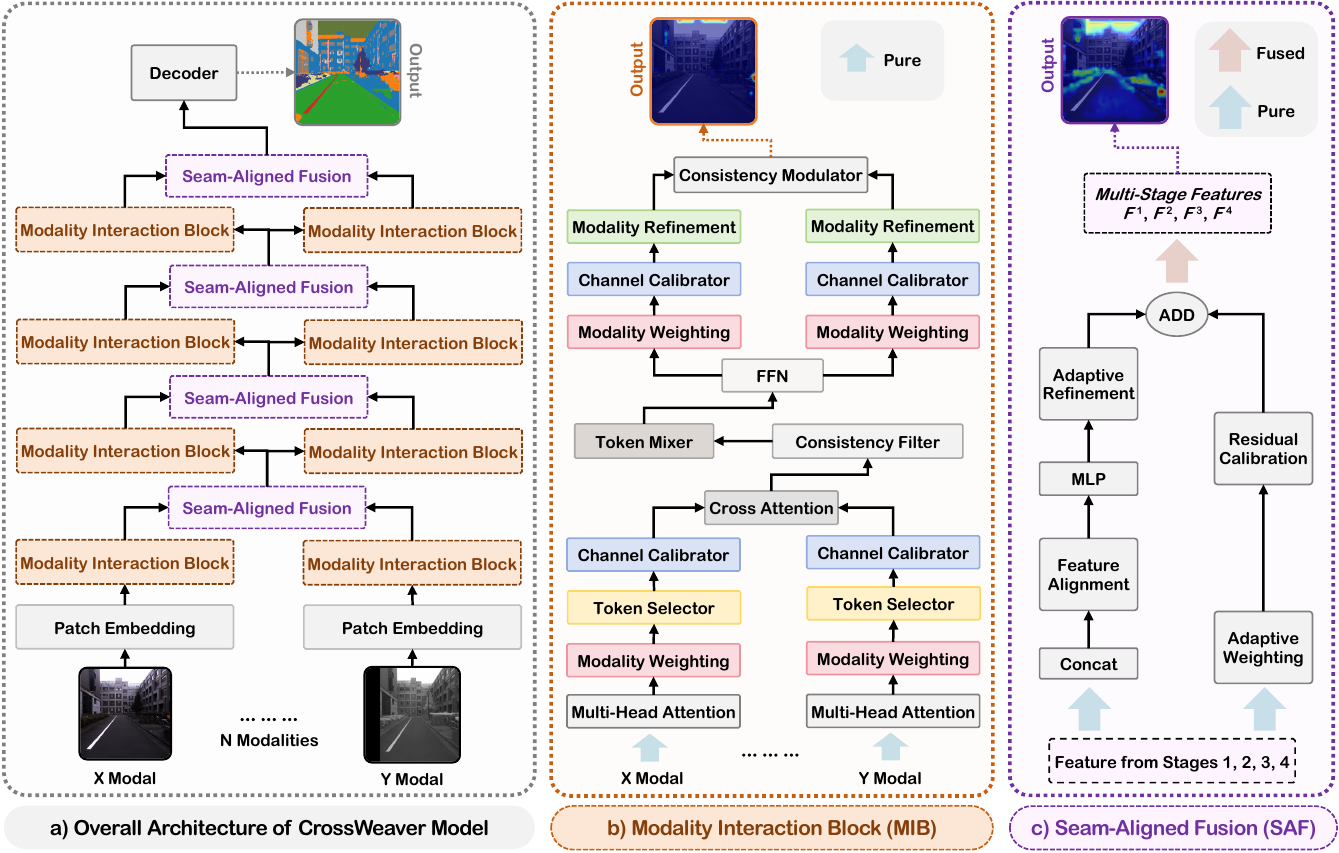}
  \caption{
  \textbf{Overall framework of CrossWeaver}, consisting of a shared hierarchical encoder and two plug-and-play modules: 
(a) Overall Architecture, 
(b) Modality Interaction Block (MIB) for Reliability-Aware Cross-Modal Encoding, and 
(c) Seam-Aligned Fusion (SAF) for Boundary Preserving Feature Fusion.
  }
  \label{fig:main}
\end{figure*}

We introduce CrossWeaver, a reliability-aware multimodal feature interaction framework for semantic segmentation, as shown in Fig.~\ref{fig:main}. It integrates the Modality Interaction Block (MIB) and the Seam-Aligned Fusion (SAF) modules to selectively exploit complementary information from different modalities.
\subsection{Task Parameterization}
\label{subsec:task}

\paragraph{Inputs.}
For \(m\) synchronized sensing modalities, the input is defined as:
\begin{equation}
\mathcal{I} = \{ I_i \in \mathbb{R}^{H \times W \times C_i} \mid i = 1,2,\dots,m \},
\label{eq:input}
\end{equation}
where \(H\) and \(W\) denote the spatial height and width, respectively, and \(C_i\) is the channel dimension of the \(i\)-th modality. For example, the modalities may include RGB, depth, event, and LiDAR.

CrossWeaver processes each modality with a weight-shared hierarchical encoder and produces stage-wise features. At stage \(l \in \{1,2,3,4\}\), the encoder outputs:
\begin{equation}
Z_i^{(l)} \in \mathbb{R}^{B \times N_l \times C_l}, \quad
N_l = H_l W_l,
\label{eq:feat}
\end{equation}
where \((H_l, W_l)\) and \(C_l\) are the spatial resolution and channel width of stage \(l\).

\paragraph{Outputs.}
Our goal is to predict a per-pixel semantic label for $L$ classes.
Given the fused multiple-scale feature pyramid
$\{F^{(l)}\}_{l=1}^{S}$ produced by CrossWeaver,
a lightweight decoder $H(\cdot)$ outputs the logits:
\begin{equation}
\hat{Y} = H\big([F^{(1)},F^{(2)},\dots,F^{(S)}]\big)
\in \mathbb{R}^{H \times W \times L},
\label{eq:decode}
\end{equation}
which are trained with a pixel-wise cross-entropy loss. With the task and representation format defined, we now detail the overall architecture and its two key components.

\subsection{CrossWeaver Architecture}
\label{subsec:arch}
Unlike conventional multimodal segmentation frameworks that rely on fixed fusion rules or heavy modality-specific branches \cite{hazirbas2016fusenet,park2017rdfnet,zhang2023cmx,reza2024mmsformer}, CrossWeaver employs a unified encoder fusion architecture that enables dynamic feature interaction across heterogeneous modalities.
Each modality stream shares hierarchical encoder weights to promote semantic alignment, while the proposed interaction modules are inserted into Transformer stages to enable selective information exchange across modalities.
This design allows modality cooperation to emerge naturally within the encoder, and the Seam-Aligned Fusion (SAF) module further refines the aggregated features into a unified semantic space.

As illustrated in Fig.~\ref{fig:main}, a SegFormer-style hierarchical backbone with $S{=}4$ stages extracts modality-specific tokens at multiple scales. 
Each encoder stage is implemented as a Modality Interaction Block (MIB), which extends the standard Transformer encoder layer by inserting cross-modal interaction modules after both the self-attention and feed-forward (FFN) components. 
This design enables fine-grained and reliability-aware information exchange across modalities while preserving the efficiency of the backbone. 
The interacted features $\{\tilde{Z}_i^{(l)}\}_{i=1}^{m}$ are then passed to the Seam-Aligned Fusion (SAF) module, which consolidates them into a single fused representation:
\begin{equation}
F^{(l)} = \mathrm{SAF}\big(\{\tilde{Z}_i^{(l)}\}_{i=1}^{m}\big)
\in \mathbb{R}^{B \times N_l \times C_l}.
\end{equation}
The fused pyramid $\{F^{(l)}\}_{l=1}^{4}$ is subsequently decoded by $H(\cdot)$ according to Eq.~\eqref{eq:decode}.
Since MIB and SAF are built upon a shared hierarchical backbone, CrossWeaver preserves the overall encoder-decoder structure while enhancing robustness under arbitrary modality combinations. We next detail the design of its two building blocks, starting with the Modality Interaction Block (MIB).

\subsection{Modality Interaction Block (MIB)}
\label{subsubsec:mib}

The Modality Interaction Block (MIB) serves as the core component of CrossWeaver, enabling adaptive, confidence-aware feature interaction among multiple modalities within the encoder. 
Each MIB sequentially performs intra-modal enhancement, reliability estimation, cross-modal exchange, and feed-forward refinement to achieve fine-grained and robust multimodal representation learning, as illustrated in Fig.~\ref{fig:main}(b).

Given the input feature $Z_i^{(l)}$ from the $i$-th modality at stage $l$, the block first applies SegFormer-style multi-head self-attention (MSA) with spatial reduction ratio $\mathrm{sr}$ to capture long-range dependencies within each modality:
\begin{equation}
\hat{Z}_i^{(l)} = Z_i^{(l)} + \mathrm{MSA}\!\big(\mathrm{LN}(Z_i^{(l)})\big),
\quad i=1,\dots,m ,
\label{eq:mib_intra}
\end{equation}
where $\mathrm{LN}$ denotes LayerNorm. 
To assess the reliability of each modality, a global descriptor $G_i^{(l)} = \mathrm{GAP}(\hat{Z}_i^{(l)}) \in \mathbb{R}^{B\times C_l}$ is computed, from which small MLPs generate both a modality-level weight $w_i^{(l)}$ and a token coverage ratio $p_i^{(l)}$:
\begin{equation}
w_i^{(l)} = \mathrm{softmax}_i\!\big(f(G_i^{(l)})/\tau_m\big), \quad
p_i^{(l)} = \sigma\!\big(h(G_i^{(l)})\big).
\label{eq:mib_weights}
\end{equation}
The token selector then predicts token-wise confidence 
$s_i^{(l)} = \sigma(\phi(\hat{Z}_i^{(l)})) \in (0,1)^{B\times N_l}$, 
from which a soft Top-$p$ mask is derived to retain informative regions:
\begin{equation}
a_i^{(l)} = \sigma\!\left(\frac{s_i^{(l)} - \theta_i^{(l)}}{\tau_a}\right), 
\quad \theta_i^{(l)}=\text{$p_i^{(l)}$-quantile}(s_i^{(l)}),
\label{eq:mib_mask}
\end{equation}
where $\tau_a$ controls mask softness. 
The channel calibrator further rescales the features, yielding reliability-filtered sources:
\begin{equation}
\breve{Z}_i^{(l)} =
\mathrm{Calib}(\hat{Z}_i^{(l)}) \odot a_i^{(l)} \odot w_i^{(l)} .
\label{eq:mib_source}
\end{equation}

Following modality-specific calibration, the MIB enters its core stage of interaction across modalities and consistency filtering.
For each ordered pair $(i\!\leftarrow\!j)$, queries are taken from $\hat{Z}_i^{(l)}$ and keys/values from $\breve{Z}_j^{(l)}$. 
The source tokens are spatially pooled into multiple grids $\{(G_h^{(s)},G_w^{(s)})\}_{s=1}^{S_g}$ to capture multi-scale dependencies. 
At each scale $s$, cross-attention is computed with a Gaussian relative positional bias $\Delta^{(s)}$ to model geometric alignment:
\begin{equation}
Y^{(s)}_{i\leftarrow j} =
\mathrm{softmax}\!\left(
\frac{Q_i K^{(s)\!\top}_j}{\sqrt{d}} + \Delta^{(s)}
\right)V^{(s)}_{j},
\label{eq:mib_xattn}
\end{equation}
where $d$ is the attention head dimension. 
The multiple scale responses are aggregated via source dependent coefficients $\alpha_j^{(l,s)}$:
\begin{equation}
\bar{Y}_{i\leftarrow j} =
\sum_{s=1}^{S_g} \alpha_j^{(l,s)}\, Y^{(s)}_{i\leftarrow j}.
\label{eq:mib_multiscale}
\end{equation}
To suppress inconsistent or noisy cross-modal messages, a cosine-similarity modulation, called the consistency filter, is applied:
\begin{equation}
\tilde{Y}_{i\leftarrow j} =
\sigma\!\big(\kappa \cdot (\hat{Z}_i^{(l)}, \bar{Y}_{i\leftarrow j})\big)
\odot \bar{Y}_{i\leftarrow j},
\label{eq:mib_consistency}
\end{equation}
where $\kappa$ controls the modulation intensity.
Here $(\cdot,\cdot)$ denotes the cosine similarity between L2-normalized feature vectors.

Subsequently, the filtered information is selectively integrated by a token mixer, which learns token-wise mixture coefficients instead of uniformly summing all messages:
\begin{equation}
\Pi_{i\leftarrow j} =
\mathrm{softmax}_{j\neq i}\!\left(
\frac{Q_i^{\mathrm{mix}}K_j^{\mathrm{mix}}}{\tau_{\mathrm{mix}}}
\right),
\label{eq:mib_mixer}
\end{equation}
where $\tau_{\mathrm{mix}}$ regulates mixture smoothness. 
The aggregated update is then computed and added back to the target modality through a residual connection:
\begin{equation}
\Delta_i^{(l)}=\sum_{j\neq i}\Pi_{i\leftarrow j}\odot \tilde{Y}_{i\leftarrow j},
\qquad
X_i^{(l)}=\hat{Z}_i^{(l)}+\Delta_i^{(l)}.
\label{eq:mib_residual}
\end{equation}

Finally, a feed-forward refinement stage further stabilizes the output.
A lightweight FFN followed by a modality refinement module, implemented with a source-conditioned affine (SCA) adapter, provides an additional softer cross-modal refinement path:
\begin{align}
U_i^{(l)} &= X_i^{(l)} +
\mathrm{FFN}\big(\mathrm{LN}(X_i^{(l)})\big), \\
\tilde{Z}_i^{(l)} &= U_i^{(l)} +
\sum_{j\neq i}\Pi^{(2)}_{i\leftarrow j}\odot
\mathrm{SCA}_{j\rightarrow i}(U_i^{(l)}).
\end{align}
Here, $\mathrm{SCA}_{j\rightarrow i}$ predicts channel wise scale and shift from the source modality $j$, 
and its residual coefficient is initialized to $2{\times}10^{-2}$ to maintain identity behavior at the start of training.
Overall, the MIB follows a progressive pipeline of self attention, modality weighting, token selection, channel calibration, cross attention, consistency filtering, token mixing, and modality refinement, as illustrated in Fig.~\ref{fig:main}(b).
Through this process, MIB achieves dynamic, reliability-aware feature interaction across modalities while preserving computational efficiency.

Compared with prior multimodal transformers that perform one-shot concatenation or late fusion, MIB emphasizes progressive reliability-aware message passing.
By combining modality confidence estimation, masked token selection, and consistency filtering, MIB dynamically suppresses corrupted or redundant sources while reinforcing informative ones. This selective interaction mechanism allows CrossWeaver to generalize across unseen modality combinations, which is particularly valuable in real-world multimodal perception where sensor failures and degradations frequently occur.

\subsection{Seam-Aligned Fusion (SAF)}
\label{subsubsec:saf}

Given the interacted features $\{\tilde{Z}_i^{(l)}\}_{i=1}^{m}$, the Seam-Aligned Fusion (SAF) module produces a single fused representation that aligns object boundaries while preserving both shared semantics and modality-specific details. As shown in Fig.~\ref{fig:main}(c), tokens from all modalities are concatenated and linearly projected back to $C_l$ channels, followed by a compact multi-branch depthwise convolutional mixing operation (e.g., $3{\times}3$, $5{\times}5$, $7{\times}7$) and channel attention to enhance local consistency around modality seams:
\begin{equation}
\hat{F}^{(l)} =
\mathrm{Mix}\!\Big(
\mathrm{Lin}\big([\tilde{Z}_1^{(l)}\!\|\!\cdots\!\|\!\tilde{Z}_m^{(l)}]\big)
\Big),
\label{eq:saf_mix}
\end{equation}
where $\mathrm{Lin}(\cdot)$ and $\mathrm{Mix}(\cdot)$ denote the linear projection and spatial mixing operations, respectively. 
To keep the fusion lightweight and retain modality-specific cues, SAF further applies a modality-weighted residual connection:
\begin{equation}
\begin{split}
F^{(l)} &= \hat{F}^{(l)} + \sum_{i=1}^{m} \omega_i^{(l)}\,\tilde{Z}_i^{(l)}, \\
\omega^{(l)} &= \mathrm{softmax}\!\big(\gamma^{(l)}/\tau^{(l)}\big),
\end{split}
\label{eq:saf_residual2}
\end{equation}
where $\omega_i^{(l)}$ are learned per stage weights derived from global summaries and $\tau^{(l)}$ is a temperature parameter. 
This residual pathway adaptively balances fused and modality-specific information, improving boundary alignment without duplicating encoder computation. 
The fused features $\{F^{(l)}\}_{l=1}^{4}$ are finally reshaped and fed into the decoder as in Eq.~\eqref{eq:decode}.

\section{Experiments and Results}
\label{Experimental Results}
\subsection{Datasets}
\label{Datasets and implementation details}

We evaluate CrossWeaver on three multimodal semantic segmentation benchmarks: MCubeS \cite{liang2022mcubes}, DeLiVER \cite{zhang2023cmnext}, and MUSES \cite{muses}. These datasets cover diverse segmentation scenarios and sensor configurations. More details about dataset splits, semantic categories, input resolutions, and modality configurations are provided in Appendix Dataset Details. We use compact modality abbreviations throughout the experiments. For MCubeS, R, A, D, and N denote Image, AoLP, DoLP, and NIR, respectively. For DeLiVER, R, D, E, and L denote RGB, Depth, Event, and LiDAR, respectively. For MUSES, I, E, and L denote Image, Event, and LiDAR, respectively.

\subsection{Implementation Details} We train the proposed framework on 8 4090 NVIDIA GPUs, using AdamW optimizer ($\epsilon = 10^{-8}$, weight decay $= 10^{-2}$) with a batch size of 4 per GPU. The learning rate starts at $6\times10^{-5}$ and follows a polynomial decay across 200 epochs, 
with the first 10 epochs used for warm-up at $0.1\times$ the base rate. 
For data augmentation, we apply random resizing (scale 0.5--2.0), horizontal flipping, color jitter, Gaussian blur, and random cropping to $1024\times1024$ for DeLiVER and $512\times512$ for MCubeS. Each method utilizes the SegFormer-B0 backbone, initialized from ImageNet-1K pre-training to ensure fair comparison.
During training, all available modalities of each dataset are jointly used at every iteration. For arbitrary-modality evaluation, we simulate missing sensors at inference time by retaining only a selected modality subset and replacing the unavailable modalities with zero tensors of the same shape. This keeps the input interface identical to the full-modality model and avoids modifying the network architecture for different modality combinations. All compared methods are evaluated under the same modality-subset protocol when reproduced in our experiments.

\begin{table}[t]
  \centering
  \caption{Comparison with state-of-the-art methods on MCubeS \cite{liang2022mcubes}, DeLiVER \cite{zhang2023cmnext}, and MUSES \cite{muses} for multimodal semantic segmentation. Following common evaluation settings, MCubeS and DeLiVER are evaluated with four modalities, while MUSES is evaluated with three modalities. All methods use the MiT-B0 backbone, and all values are reported in mIoU (\%). ``--'' denotes unavailable results.}
  \label{tab:results-main}
  \resizebox{0.8\linewidth}{!}{
  \begin{tabular}{lccc}
    \toprule
    Method & MCubeS & DeLiVER & MUSES \\
    \midrule
    CMNeXt \cite{zhang2023cmnext} 
      & 40.94 & 59.84 & 46.66 \\
      
    MAGIC \cite{zheng2024centering}
      & 47.85 & 63.40 & 49.05 \\
      
    Any2Seg \cite{any2seg}
      & -- & -- & 50.00 \\
      
    AnySeg \cite{anyseg}
      & -- & 59.72 & 51.25 \\
      
    MMSFormer \cite{reza2024mmsformer} 
      & 45.98 & 58.63 & -- \\

    CHARM \cite{charm}
      & 48.24 & 64.03 & 52.31 \\
      
    StitchFusion \cite{li2025stitchfusion} 
      & 46.99 & 62.67 & -- \\
      
    \midrule
    \rowcolor{gray!20}
    CrossWeaver (Ours)
      & \textbf{48.76} & \textbf{64.69} & \textbf{53.19} \\
    \bottomrule
  \end{tabular}}
\end{table}

\begin{table*}[t]
\centering
\caption{Validation with arbitrary modality combinations after four-modality training on MCubeS \cite{liang2022mcubes} and DeLiVER \cite{zhang2023cmnext}. We include the latest reproducible state-of-the-art baselines with publicly available code and compatible protocols. All methods are trained with the MiT-B0 backbone, and all values are reported in mIoU (\%).}
\label{tab:validation-arbitrary-combinations}
\scriptsize
\setlength{\tabcolsep}{4.0pt}
\renewcommand{\arraystretch}{0.92}

\begin{tabular*}{\textwidth}{@{\extracolsep{\fill}}lcccccccccccccccc@{}}
\toprule
\multicolumn{17}{c}{\textbf{MCubeS} \cite{liang2022mcubes}} \\
\cmidrule(lr){1-17}
Method & R & A & D & N & RA & RD & RN & AD & AN & DN & RAD & RAN & RDN & ADN & RADN & Mean \\
\midrule
CMNeXt \cite{zhang2023cmnext}
& 2.53 & 1.71 & 1.28 & 0.23 & 6.47 & 7.73 & 7.56 & 4.13
& 3.76 & 0.36 & 5.37 & 5.58 & 9.14 & 4.56 & 36.16 & 6.44 \\

StitchFusion \cite{li2025stitchfusion}
& 36.77 & 3.10 & 4.41 & 12.18 & 39.61 & 41.55 & 39.30 & 8.09
& 18.76 & 20.22 & 43.91 & 42.73 & 44.60 & 25.92 & 46.99 & 28.54 \\

\rowcolor{gray!20}
CrossWeaver (Ours)
& \textbf{39.77} & \textbf{15.18} & \textbf{12.18} & \textbf{23.76}
& \textbf{41.57} & \textbf{41.92} & \textbf{42.90} & \textbf{19.39}
& \textbf{22.05} & \textbf{22.83} & \textbf{45.62} & \textbf{45.28}
& \textbf{46.05} & \textbf{23.01} & \textbf{48.76} & \textbf{32.68} \\
\midrule
\multicolumn{17}{c}{\textbf{DeLiVER} \cite{zhang2023cmnext}} \\
\cmidrule(lr){1-17}
Method & R & D & E & L & RD & RE & RL & DE & DL & EL & RDE & RDL & REL & DEL & RDEL & Mean \\
\midrule
CMNeXt \cite{zhang2023cmnext}
& 0.86 & 0.49 & 0.66 & 0.37 & 47.06 & 9.97 & 13.75 & 2.63
& 1.73 & 2.85 & 59.03 & 59.18 & 14.73 & 39.07 & 59.18 & 20.77 \\

StitchFusion \cite{li2025stitchfusion}
& 24.88 & 47.79 & 2.04 & 2.32 & 60.88 & 25.53 & 27.39 & 48.24
& 48.38 & 2.74 & 60.75 & 60.50 & 27.86 & 48.86 & 62.67 & 36.72 \\

\rowcolor{gray!20}
CrossWeaver (Ours)
& \textbf{29.92} & \textbf{53.60} & \textbf{2.53} & \textbf{4.40}
& \textbf{63.01} & \textbf{31.77} & \textbf{32.67} & \textbf{53.23}
& \textbf{53.36} & \textbf{5.49} & \textbf{62.98} & \textbf{62.64}
& \textbf{33.28} & \textbf{54.21} & \textbf{64.69} & \textbf{40.52} \\
\bottomrule
\end{tabular*}
\end{table*}

\subsection{Experimental Results}
Table~\ref{tab:results-main} compares CrossWeaver with representative state-of-the-art methods on MCubeS, DeLiVER, and MUSES. With the MiT-B0 backbone, CrossWeaver achieves 48.76\%, 64.69\%, and 53.19\% mIoU on the three datasets, respectively. On MCubeS, it improves over CHARM by +0.52\% mIoU and also outperforms MAGIC, StitchFusion, MMSFormer, and CMNeXt. On DeLiVER, CrossWeaver reaches 64.69\% mIoU, surpassing CHARM and MAGIC by +0.66\% and +1.29\%, respectively. On MUSES, it obtains 53.19\% mIoU, exceeding CHARM by +0.88\% and AnySeg by +1.94\%. These consistent gains demonstrate the effectiveness of our selective cross-modal interaction and seam-aligned fusion design across diverse multimodal segmentation benchmarks.

\paragraph{Validation with Arbitrary Modality.}
To evaluate robustness to missing modalities, we further test models trained with all available modalities on arbitrary subsets of inputs, as reported in Table~\ref{tab:validation-arbitrary-combinations}. Since this evaluation requires running each method under all modality subsets, we include the latest reproducible state-of-the-art baselines with publicly available code and compatible protocols. Recent state-of-the-art methods without released code are compared in Table~\ref{tab:results-main} using their reported results when available.

On MCubeS, CrossWeaver consistently achieves the best performance across all modality combinations, from single-modality to full-modality inputs. It reaches 39.77\% mIoU with RGB alone and maintains clear advantages under incomplete settings such as RA (41.57\%), RN (42.90\%), RAN (45.28\%), and RDN (46.05\%), while also achieving the highest score under the full RADN setting (48.76\%). Overall, its mean mIoU reaches 32.68\%, exceeding CMNeXt and StitchFusion by 26.24 and 4.14 points, respectively. These results indicate that CrossWeaver preserves strong predictive ability even when only partial modalities are available.

A similar trend is observed on DeLiVER. CrossWeaver again achieves the best overall performance across reduced and full-modality settings. It attains 29.92\% mIoU with RGB alone and 53.60\% with Depth alone, while maintaining strong results under partial inputs, including 63.01\% for RD, 62.98\% for RDE, 62.64\% for RDL, and 54.21\% for DEL. Under the full RDEL input, it reaches 64.69\% mIoU. Its mean mIoU is 40.52\%, surpassing CMNeXt and StitchFusion by 19.75 and 3.80 points, respectively. Overall, these results show that CrossWeaver generalizes effectively from full-modality training to incomplete modality inputs, highlighting its practical robustness in real-world multimodal perception.

\subsection{Ablation Studies}
\label{ablation}
\paragraph{Ablation of the CrossWeaver Architecture.}
As shown in Table \ref{tab:ablation-cw}, we ablate the key components of the CrossWeaver framework to assess their individual contributions.
When removing the Modality Interaction Block (MIB), the mIoU drops by 4.37\%, indicating that dynamic token-level interaction between modalities plays a crucial role in capturing complementary cues.
Eliminating the Seam-Aligned Fusion (SAF) module leads to a 1.35\% decrease, which shows that SAF effectively refines cross-modal aggregation by preserving boundary consistency and reducing feature misalignment.
When both modules are removed, the performance significantly decreases to 41.74\% mIoU, highlighting that MIB and SAF are jointly responsible for robust multimodal feature fusion and fine-grained semantic reasoning.
Overall, these results verify that both interaction and fusion mechanisms are essential for achieving strong multimodal segmentation performance in CrossWeaver.

\begin{table}[t]
  \centering
  \caption{Ablation Study of the CrossWeaver Architecture.}
  \label{tab:ablation-cw}
  \small
  \resizebox{0.7\linewidth}{!}{
  \begin{tabular}{ccccc}
    \toprule
    \textbf{MIB} & \textbf{SAF} & \textbf{\#Params (M)} & \textbf{GFLOPs} & \textbf{mIoU (\%)} \\
    \midrule
    $\checkmark$ & $\checkmark$ & 12.31 & 37.68 & 48.76 \\
    $\times$     & $\checkmark$ & 11.48 & 33.85 & 47.41 \\
    $\checkmark$ & $\times$     & 6.55  & 29.34 & 44.39 \\
    $\times$     & $\times$     & 3.72  & 26.72 & 41.74 \\
    \bottomrule
  \end{tabular}
  }
\end{table}

\begin{figure*}[t]
  \centering
  \includegraphics[width=\linewidth]{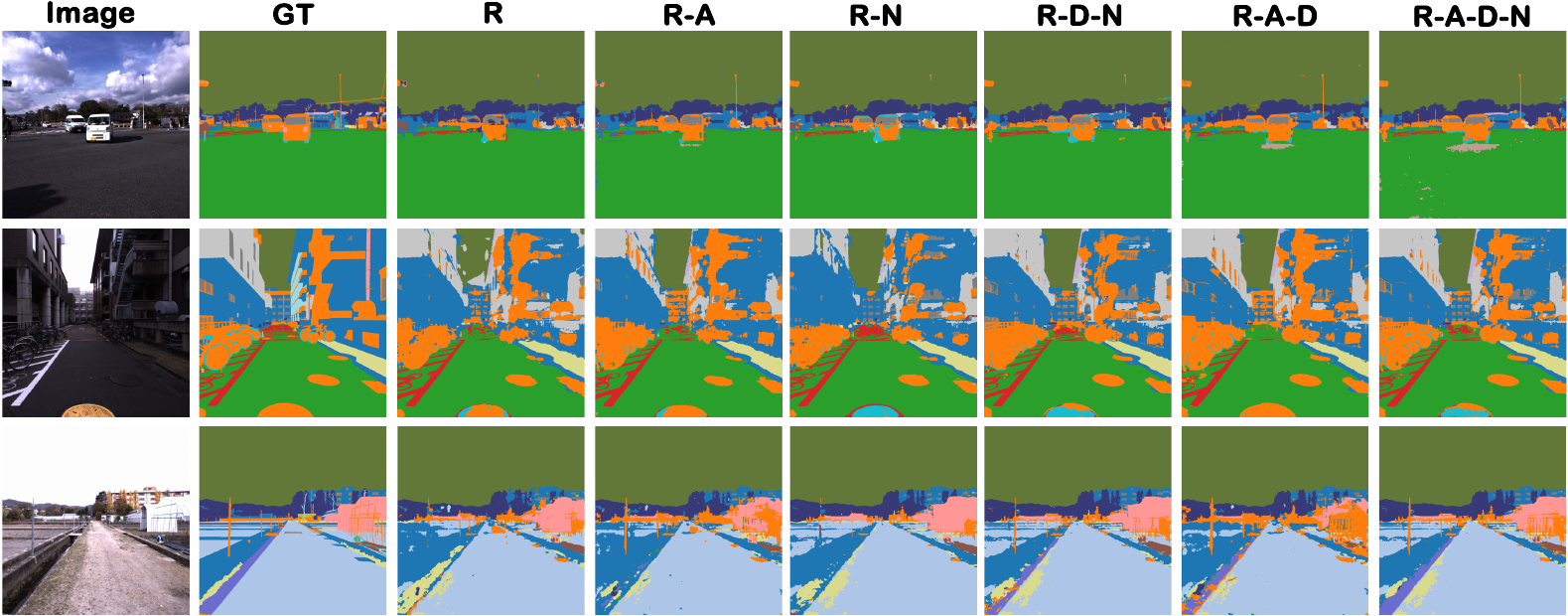}
  \caption{Visualization of CrossWeaver (MiT-B0 Backbone) on the MCubeS \cite{liang2022mcubes} Dataset.}
  \label{Visiualization}
\end{figure*}

\begin{figure}
  \centering
  \includegraphics[width=\linewidth]{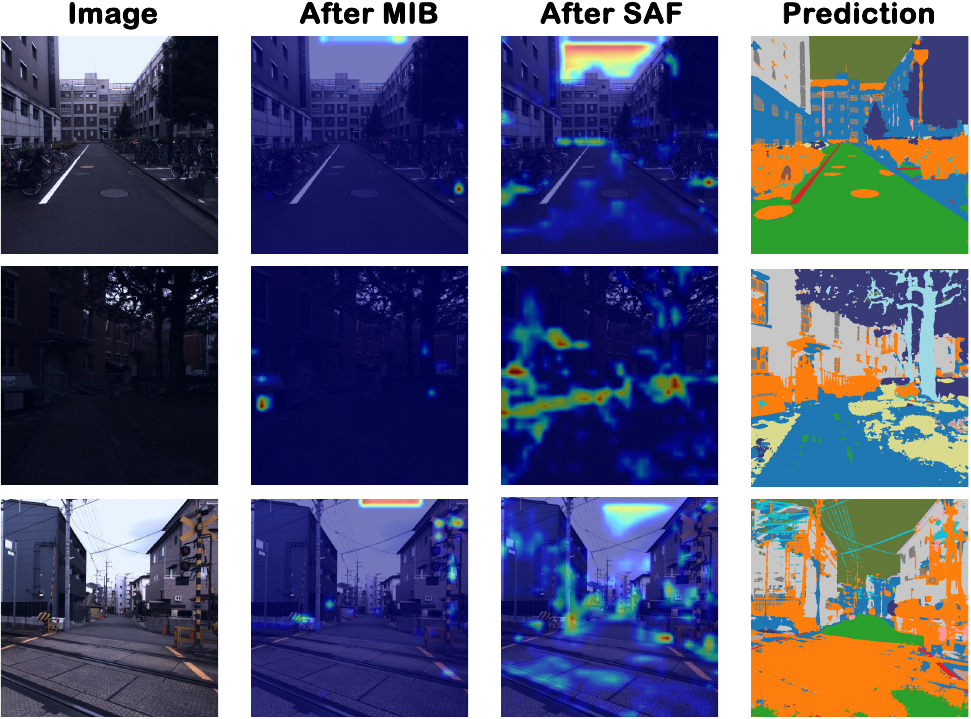}
  \caption{Qualitative Visualization of MIB and SAF in CrossWeaver on the MCubeS Dataset. From left to right are the input image, the feature response after MIB, the feature response after SAF, and the final prediction.}
  \label{fig:module_vis}
  \vspace{-10pt}
\end{figure}

\paragraph{Effect of MIB Insertion Stage.}
To analyze the impact of inserting the Modality Interaction Block (MIB) at different stages, we conduct a stage-wise ablation on MCubeS with the MiT-B0 backbone, as shown in Table~\ref{tab:mib-stage}.
When MIB is applied at early stages, it mainly enhances low-level feature consistency (e.g., texture and polarization cues), 
yielding moderate gains. As it is progressively inserted into deeper stages, performance improves steadily due to 
richer semantic interactions across modalities. The best results are obtained when MIB is deployed across all stages (stage 0–4), 
achieving 48.76\% mIoU and 58.78\% pixel accuracy, indicating that both early structural alignment and high-level semantic fusion are crucial for comprehensive multimodal understanding.

\paragraph{Semantic Segmentation Visualization.}
Fig.~\ref{Visiualization} presents qualitative segmentation results of CrossWeaver (MiT-B0) on the MCubeS dataset under different modality combinations. 
The model is trained with four modalities and evaluated with progressively richer inputs, including R, R--A, R--A--D, and R--A--D--N. 
For reference, we also provide the input image and ground-truth annotation in each row. 
As more modalities are introduced, the predicted segmentation maps become closer to the ground truth and exhibit clearer structures, more complete object regions, and better boundary delineation. 
Notably, CrossWeaver still produces reasonable predictions under reduced modality settings, while additional modalities further improve semantic completeness and spatial consistency. 
These results qualitatively demonstrate both the effectiveness of complementary multimodal fusion and the robustness of CrossWeaver under missing-modality conditions.

\paragraph{Qualitative Analysis of MIB and SAF.}
To provide a more intuitive understanding of the roles of the two proposed modules, we further visualize the intermediate feature responses. As shown in Fig.~\ref{fig:module_vis}, the feature map after MIB already exhibits a certain degree of selectivity, indicating that modality interaction helps the network focus on more informative regions. After SAF, the fused representation becomes more spatially coherent, suggesting that this module facilitates more consistent cross-modal integration and refinement. These qualitative observations are consistent with the ablation results in Table~\ref{tab:ablation-cw}, and provide additional support for the effectiveness of both modules.



\begin{table}[t]
  \centering
  \caption{Ablation of MIB Insertion across Encoder Stages on MCubeS Dataset (MiT-B0 Backbone). 
  "Stage-X" means that MIB is applied at the $X$-th encoder stage.}
  \label{tab:mib-stage}
  \resizebox{0.9\linewidth}{!}{
  \begin{tabular}{lccccc}
    \toprule
    Metric & Stage1 & Stage2 & Stage3 & Stage4 & Stage1--4 \\
    \midrule
    mIoU (\%) & 47.16 & 46.91 & 47.47 & 46.65 & \textbf{48.76} \\
    Acc (\%)  & 56.93 & 56.53 & 57.14 & 56.44 & \textbf{58.78} \\
    \bottomrule
  \end{tabular}
  }
\end{table}

\section{Conclusion}
\label{conclusion}

In this paper, we presented CrossWeaver, a reliability-aware multimodal interaction framework for arbitrary-modality semantic segmentation. With the Modality Interaction Block (MIB) and Seam-Aligned Fusion (SAF), CrossWeaver selectively exchanges informative cross-modal cues and aggregates them with improved spatial coherence. Experiments on MCubeS, DeLiVER, and MUSES show strong performance against recent multimodal and arbitrary-modality methods, while ablation studies verify the effectiveness of the proposed components. The current design still assumes spatially registered inputs and introduces additional computation over a plain shared encoder. Future work will explore more efficient interaction mechanisms and extensions to dynamic or strongly misaligned multimodal scenarios.


\clearpage
{
    \small
    \bibliographystyle{ieeenat_fullname}
    \bibliography{reference}
}

\clearpage
\appendix
\twocolumn[
\begin{@twocolumnfalse}
\begin{center}
    {\LARGE \textbf{Technical Appendices and Supplementary Material}}
\end{center}
\vspace{1em}
\end{@twocolumnfalse}
]
This supplementary document provides additional experimental results to further validate the effectiveness and robustness of CrossWeaver. We first present comprehensive qualitative visualizations on MCubeS and DeLiVER under varying modality combinations to examine the model’s behavior when partial sensory inputs are available. We then illustrate segmentation performance on DeLiVER across diverse adverse weather conditions to evaluate real-world deployment reliability. Together, these results demonstrate that CrossWeaver maintains stable semantic understanding, gracefully degrades under missing or unreliable modalities, and consistently leverages complementary sensor cues to preserve scene structure and object boundaries.

\section{Dataset Details}
\label{sec:dataset-details}

\paragraph{MCubeS.}
MCubeS is a multimodal material segmentation benchmark with 20 semantic classes. It provides co-registered RGB, Near-Infrared (NIR), Degree of Linear Polarization (DoLP), and Angle of Linear Polarization (AoLP) images. The official split contains 302 training, 96 validation, and 102 test samples, each at a resolution of $1224{\times}1024$.

\paragraph{DeLiVER.}
DeLiVER is a large-scale multimodal semantic segmentation dataset built on the CARLA simulator. It offers time-synchronized RGB, Depth, Event, and LiDAR streams, together with multi-view imagery arranged as a panoramic cubemap. The dataset spans diverse weather and lighting conditions, including clear, cloudy, foggy, rainy, and nighttime scenes, and introduces sensor impairments such as motion blur, exposure variation, and LiDAR jitter. In total, it comprises 47{,}310 frames with pixel-wise semantic and instance annotations across 25 fine-grained categories, targeting robust multimodal perception under challenging conditions.

\paragraph{MUSES.}
MUSES is a real-world multi-sensor semantic perception dataset designed for driving under uncertainty. It contains synchronized sensory data collected from diverse urban driving scenarios, with pixel-level semantic annotations for evaluating robust scene understanding under challenging conditions. Following recent modality-agnostic semantic segmentation settings, we use the RGB, Event, and LiDAR modalities for evaluation. This dataset complements MCubeS and DeLiVER by providing real-world driving scenes with heterogeneous sensor inputs, allowing us to further examine the generalization ability of CrossWeaver across different multimodal perception scenarios.

\section{Additional Quantitative Results}
\label{sec:supp-quantitative}

\begin{figure*}[t]
  \centering
  \includegraphics[width=\linewidth]{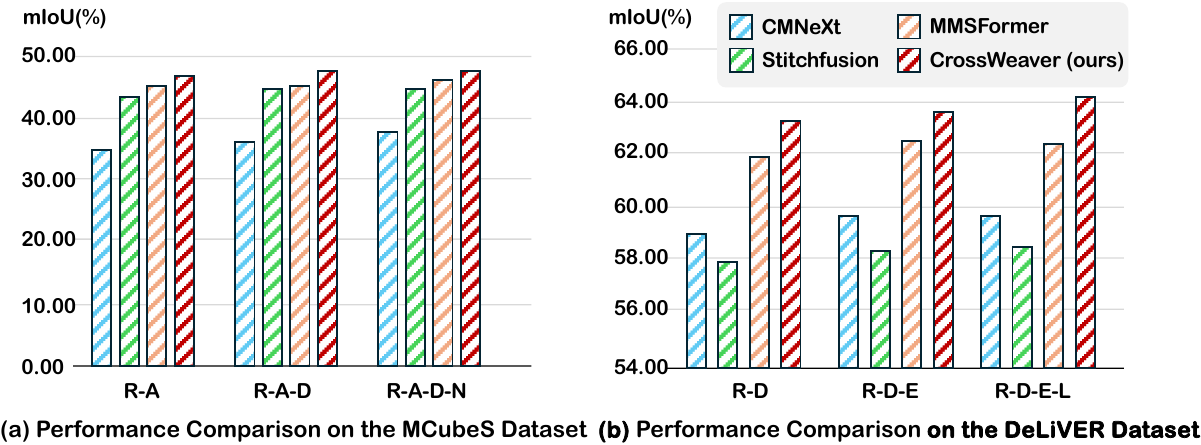}
  \caption{
  Supplementary comparison with state-of-the-art methods on MCubeS and DeLiVER under different modality settings. The left panel reports results on MCubeS, and the right panel reports results on DeLiVER.
  }
  \label{fig:supp-comparison}
\end{figure*}

Fig.~\ref{fig:supp-comparison} and Table~\ref{tab:supp-modality-results} provide additional quantitative comparisons on MCubeS and DeLiVER under separately trained modality settings. Specifically, each model is trained and evaluated with the same listed modality combination, such as R--A, R--A--D, or R--A--D--N on MCubeS, and R--D, R--D--E, or R--D--E--L on DeLiVER. This setting differs from the arbitrary-modality evaluation in main table, where models are trained with all available modalities and tested under different modality subsets.

\paragraph{Comparison Scope.}
For the arbitrary-modality evaluation in main table, we include reproducible baselines for which public code and compatible evaluation protocols are available. This evaluation requires running each method under all modality subsets after full-modality training, and therefore cannot be reliably obtained from reported full-modality numbers alone. Some recent state-of-the-art methods are included in the main comparison table using their reported results when available, but are not included in the arbitrary-modality table because executable official implementations, checkpoints, or sufficiently specified reproduction settings were not available to us at the time of evaluation. Including such methods through unofficial re-implementations or directly copying non-matched reported numbers would mix training data, preprocessing, and evaluation protocols. We therefore restrict the arbitrary-modality comparison to reproducible methods under a consistent protocol.

\paragraph{Results on MCubeS.}
On the MCubeS dataset, CrossWeaver achieves the best performance across all evaluated modality settings. Specifically, it obtains 47.92\%, 48.20\%, and 48.76\% mIoU under R--A, R--A--D, and R--A--D--N, respectively, outperforming representative baselines including CMNeXt, MMSFormer, and StitchFusion. These results show that CrossWeaver benefits from richer modality combinations while maintaining strong performance under reduced inputs.

\paragraph{Results on DeLiVER.}
On the DeLiVER dataset, CrossWeaver also consistently achieves the highest mIoU across the evaluated modality settings. It reaches 63.34\%, 63.71\%, and 64.69\% mIoU under R--D, R--D--E, and R--D--E--L, respectively. The consistent improvements over CMNeXt, MMSFormer, and StitchFusion indicate that CrossWeaver can effectively leverage complementary RGB, Depth, Event, and LiDAR cues for robust semantic segmentation.

\begin{table}
  \centering
  \caption{Supplementary comparison with state-of-the-art methods on the MCubeS and DeLiVER datasets for multimodal semantic segmentation. All methods are trained with the MiT-B0 backbone. All values are reported in mIoU (\%). Abbreviations: for MCubeS, R = Image, A = AoLP, D = DoLP, N = NIR; for DeLiVER, R = RGB, D = Depth, E = Event, L = LiDAR.}
  \label{tab:supp-modality-results}
  \resizebox{\linewidth}{!}{
  \begin{tabular}{lcccccc}
    \toprule
    \multirow{2}{*}{Method} & \multicolumn{3}{c}{MCubeS} & \multicolumn{3}{c}{DeLiVER} \\
    \cmidrule(lr){2-4} \cmidrule(lr){5-7}
    & R-A & R-A-D & R-A-D-N & R-D & R-D-E & R-D-E-L \\
    \midrule
    CMNeXt 
      & 36.16 & 37.21 & 40.94 & 59.18 & 59.61 & 59.84 \\
    MMSFormer 
      & 44.79 & 45.38 & 45.98 & 58.14 & 58.37 & 58.63 \\
    StitchFusion 
      & 46.01 & 46.22 & 46.99 & 61.98 & 62.23 & 62.67 \\
    \rowcolor{gray!20} CrossWeaver (Ours)
      & \textbf{47.92} & \textbf{48.20} & \textbf{48.76}
      & \textbf{63.34} & \textbf{63.71} & \textbf{64.69} \\
    \bottomrule
  \end{tabular}}
\end{table}

\paragraph{Additional Ablation of the Modality Interaction Block.}
\label{sec:appendix-mib-ablation}

To further analyze the contribution of each component inside the Modality Interaction Block (MIB), we conduct a component-wise ablation study on the MCubeS dataset. Starting from the full CrossWeaver model, we remove one component at a time while keeping the remaining architecture and training protocol unchanged. The results are reported in Table~\ref{tab:appendix-mib-component-ablation}.

\begin{table}[t]
  \centering
  \caption{Additional ablation of the Modality Interaction Block (MIB) on the MCubeS dataset. All variants use the MiT-B0 backbone and are evaluated under the full-modality setting.}
  \label{tab:appendix-mib-component-ablation}
  \small
  \resizebox{0.95\linewidth}{!}{
  \begin{tabular}{lcc}
    \toprule
    Variant & mIoU (\%) & Drop \\
    \midrule
    Full CrossWeaver & \textbf{48.76} & -- \\
    w/o Modality Weighting & 48.32 & -0.44 \\
    w/o Token Selector & 48.19 & -0.57 \\
    w/o Gaussian Positional Bias & 48.29 & -0.47 \\
    w/o Consistency Filter & 48.28 & -0.48 \\
    w/o Token Mixer & 48.26 & -0.50 \\
    w/o Modality Refinement (SCA) & 48.61 & -0.15 \\
    \bottomrule
  \end{tabular}}
\end{table}

As shown in Table~\ref{tab:appendix-mib-component-ablation}, removing any individual component from MIB leads to a performance decrease, indicating that these components provide complementary benefits for reliability-aware cross-modal interaction. Removing the token selector causes the largest drop, from 48.76\% to 48.19\% mIoU, suggesting that selecting informative tokens is important for suppressing redundant or noisy regions before cross-modal exchange. Removing the token mixer also reduces performance to 48.26\% mIoU, showing that adaptive token-wise integration is more effective than directly aggregating cross-modal messages.

The removal of modality weighting, Gaussian positional bias, and consistency filtering reduces the mIoU to 48.32\%, 48.29\%, and 48.28\%, respectively. These results verify the importance of estimating modality reliability, preserving spatial correspondence during cross-attention, and filtering inconsistent cross-modal responses. Removing the modality refinement module implemented by the SCA adapter leads to a smaller but still consistent drop from 48.76\% to 48.61\% mIoU, indicating that the final refinement stage further calibrates cross-modal features after token-level interaction. Overall, the component-wise ablation confirms that MIB benefits from the joint design of modality weighting, token selection, spatially aware cross-attention, consistency filtering, token mixing, and modality refinement.

\section{Supplementary Visualization of CrossWeaver on the MCubeS dataset}
\begin{figure*}[t]
  \centering
  \includegraphics[width=\linewidth]{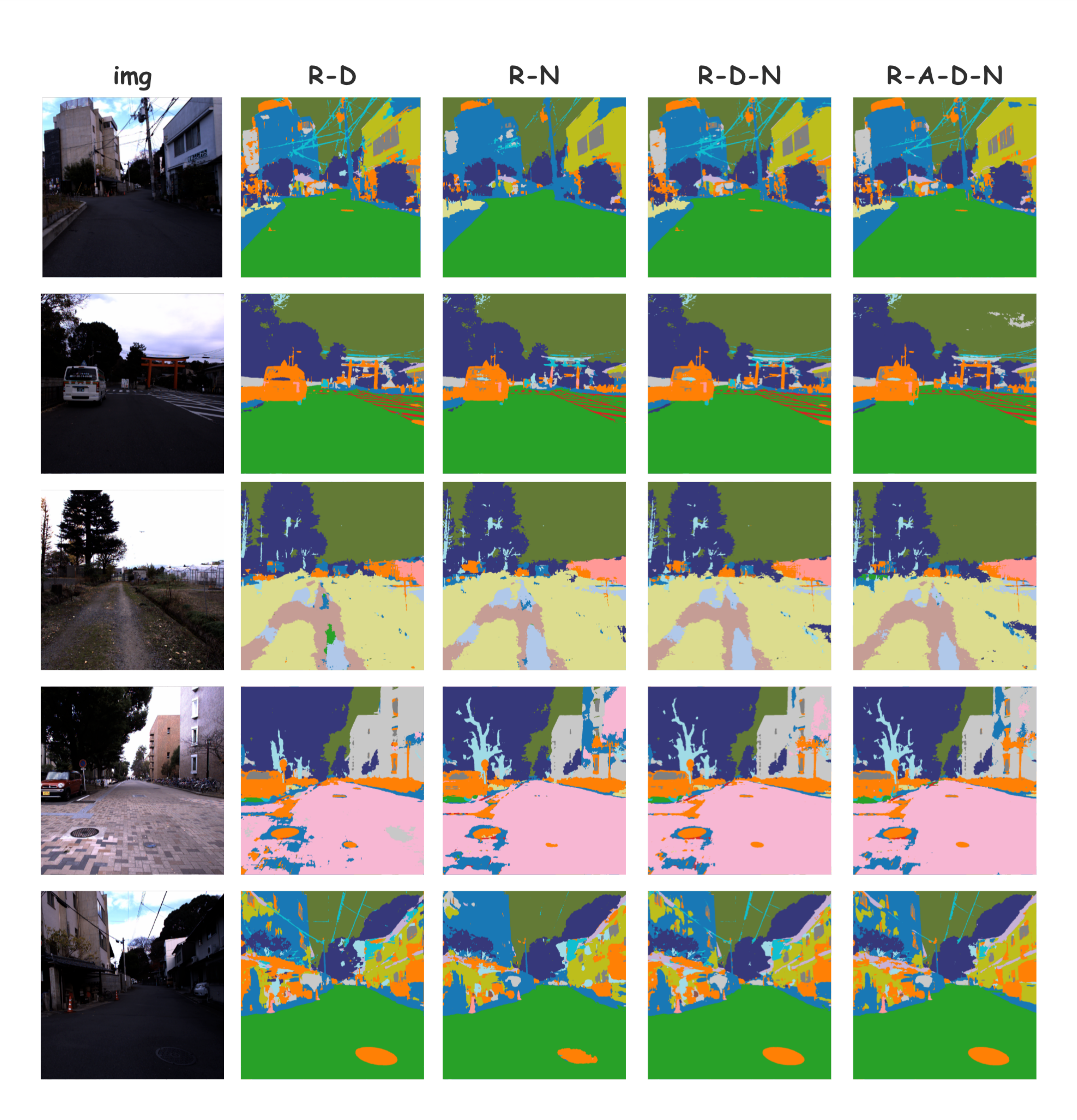}
  \caption{Additional visualizations on MCubeS under missing-modality conditions.}
  \label{fig:full-val-mcubes}
\end{figure*}

Fig.~\ref{fig:full-val-mcubes} presents additional qualitative visualizations on the MCubeS dataset, where CrossWeaver is trained with all four modalities (RGB, AoLP, DoLP, NIR) but evaluated using different modality combinations. Each column corresponds to a specific input modality set, ranging from dual-modal (R–D) to tri-modal (R–D–N) and the full-modal setting (R–A–D–N). 

Across all scenes, CrossWeaver demonstrates remarkable robustness to missing modality inputs: When switching from R–D to R–N, the model maintains stable road boundaries and structural contours, suggesting that AoLP and NIR can serve as effective substitutes when depth cues are not available. Under the R–D–N configuration (removing AoLP), the semantic predictions remain highly consistent, preserving thin structures such as poles, road curbs and building outlines — indicating that the interaction mechanism efficiently merges redundant geometric and illumination cues. The full-modality setting R–A–D–N yields the cleanest segmentation, with reduced noise on small objects and improved consistency on difficult regions such as vegetation edges and low-texture backgrounds.

These observations validate that CrossWeaver does not collapse when certain sensors are missing, but instead gracefully degrades by leveraging remaining modalities. The model consistently preserves the global layout and class-level semantics, even when informative cues (e.g., depth or polarization) are removed. This demonstrates the strong generalization ability of the proposed modality interaction and fusion strategy under unreliable or dynamically varying sensor availability, a crucial requirement in real-world robotic and autonomous perception.

\begin{figure*}[t]
  \centering
  \includegraphics[width=\linewidth]{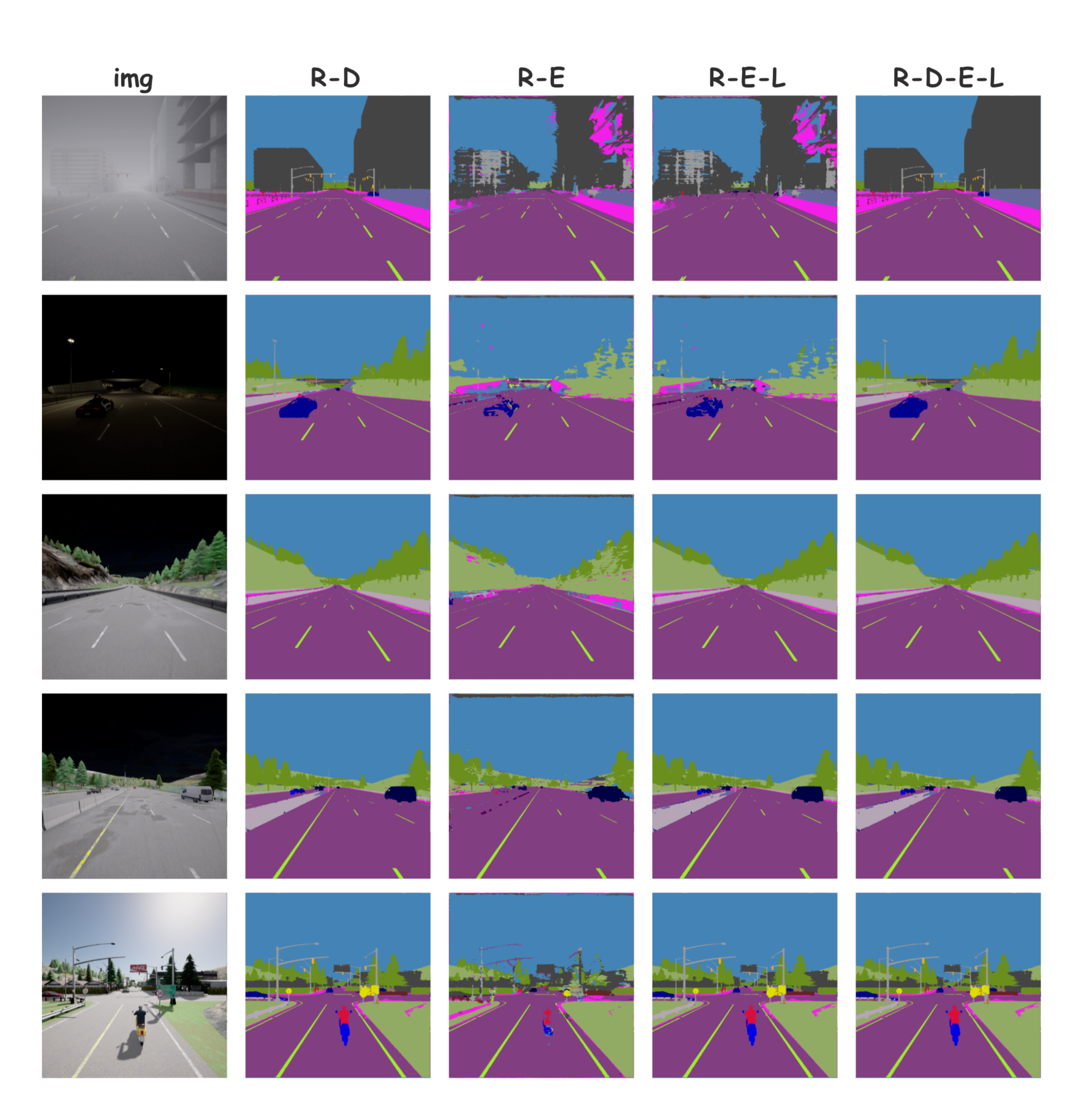}
  \caption{Additional visualizations on DeLiVER under missing-modality conditions. }
  \label{fig:fullvis-deliver}
\end{figure*}

\section{Supplementary Visualization of CrossWeaver on the DeLiVER dataset}
Fig.~\ref{fig:fullvis-deliver} presents additional qualitative visualizations on the DeLiVER dataset, where CrossWeaver is trained with all four modalities (RGB, Depth, LiDAR, Event) and evaluated with different modality subsets. This evaluation isolates the contribution of each modality and demonstrates the behavior of CrossWeaver under partial and full sensory input availability.

Across diverse driving scenes, a consistent pattern can be observed. When evaluated with only RGB, the model already recovers the coarse drivable area and major object regions, though thin structures and far-distance geometry may become unstable. Introducing Depth (R--D) substantially improves boundary sharpness on road surfaces and curbs, reflecting the strong geometric cues provided by depth sensing. When Depth is replaced by Event (R--E), the model retains clear motion boundaries and lane markings under dynamic lighting, indicating that Event signals act as an effective substitute when geometric depth is unreliable.

\begin{figure*}[t]
  \centering
  \includegraphics[width=\linewidth]{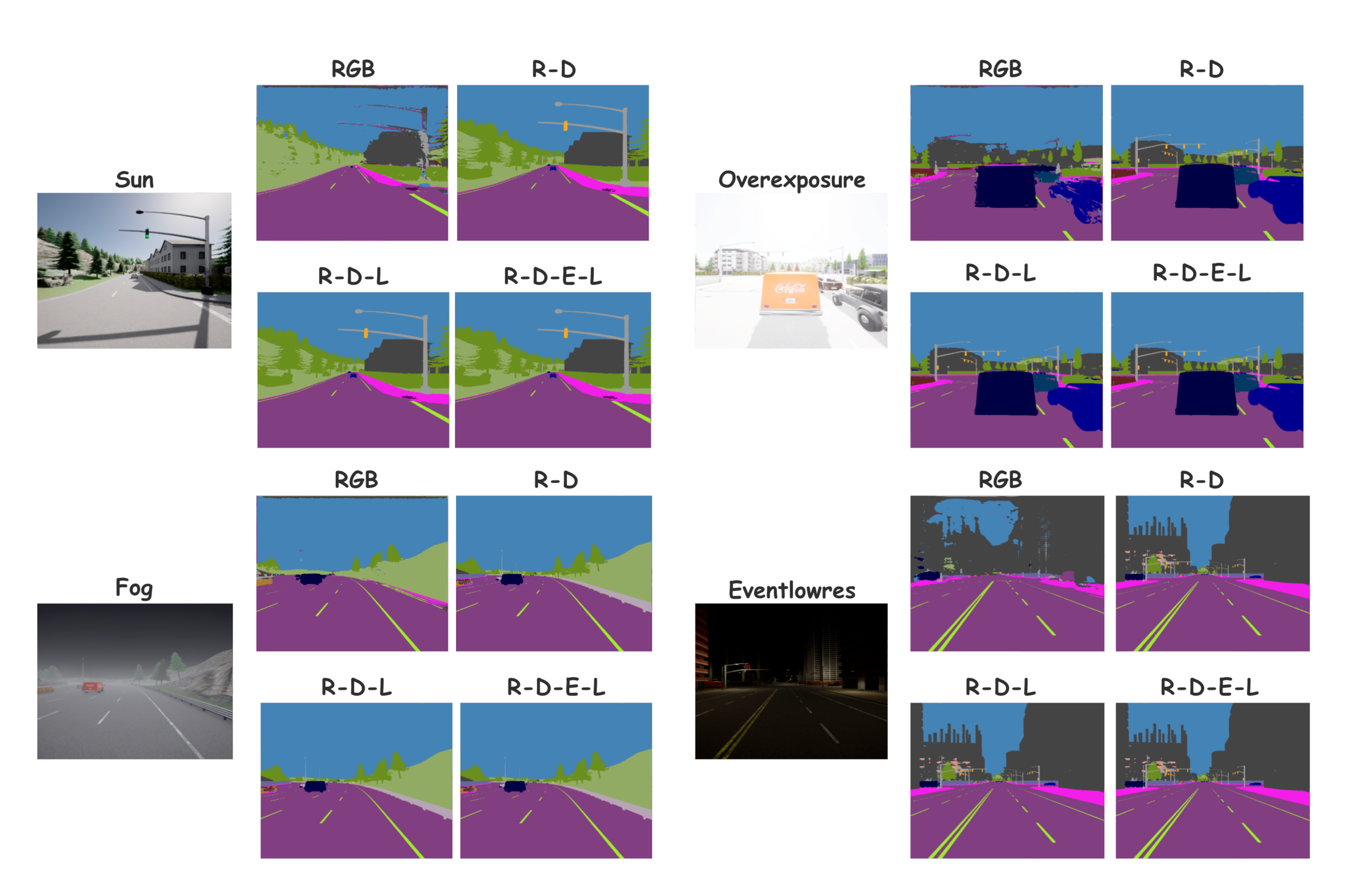}
  \caption{Additional visualizations on the DeLiVER dataset under diverse weather conditions and varying modality combinations.}
  \label{fig:weather}
\end{figure*}

Adding LiDAR (R--E--L or R--D--L) further reduces noise in cluttered regions such as vegetation, sidewalk edges, and traffic poles. These modalities collectively provide long-range geometry and high-frequency spatial structure, leading to more stable segmentations in low-texture and reflective regions. The best overall performance is achieved in the full-modality configuration (R--D--E--L), which produces the cleanest semantic maps, minimizes boundary ambiguity, and recovers small objects such as pedestrians and signposts that are easily missed by unimodal inference.

Notably, even when strong modalities are removed (e.g., without Depth or Event), the segmentation quality does not collapse. Instead, CrossWeaver gracefully degrades while preserving global scene layout and class-level semantics. This indicates that the proposed cross-modal interaction design does not overly depend on any single sensor and can effectively re-balance complementary cues when sensor failure occurs. Such robustness is important for real-world multimodal perception in safety-critical applications.

\section{Visualization under Adverse Weather and Modality Combinations}
To further evaluate the robustness of CrossWeaver under real-world complexity, we additionally visualize the segmentation results on the DeLiVER dataset using the model trained with all four modalities (RGB, Depth, LiDAR, Event). The visualizations cover multiple adverse weather conditions, including sunny, foggy, overexposure, and low-light event scenarios. They also test different modality combinations during inference, ranging from a single modality (e.g., RGB) to partial combinations (e.g., R--D or R--D--L) and the full combination (R--D--E--L).
Representative results are shown in Fig.~\ref{fig:weather}.

Across all weather conditions, several consistent observations emerge: Under clear/sunny weather, CrossWeaver produces accurate road boundary and object segmentation even using only RGB, while adding Depth and LiDAR further enhances small-object delineation (e.g., poles and sidewalk ramps). Under fog, RGB performance degrades noticeably, but the model successfully compensates when Depth and LiDAR are available. The R–D–L and full R–D–E–L combinations reliably preserve drivable area and vehicle contours. Under overexposure, Event sensors provide complementary information regarding motion boundaries. When combined (R–D–E–L), the model eliminates false positives from washed-out RGB regions and restores object shapes more faithfully. Under low-light conditions (“Eventlowres”), RGB severely fails due to lack of illumination, while Event data captures motion-induced edges that guide segmentation. The model demonstrates graceful degradation: R–D preserves lane boundaries; R–D–L and R–D–E–L reconstruct the semantic layout with high fidelity.

Overall, the visualizations indicate that CrossWeaver maintains stable semantic consistency across weather variations, and performance improves monotonically as complementary modalities become available. When some modalities fail due to adverse sensing conditions, others are adaptively leveraged to prevent error propagation. This validates the core design goal of CrossWeaver, namely consistent multimodal fusion and graceful degradation under missing or unreliable modalities.

\end{document}